\documentclass[a4paper]{cas-dc}

\usepackage[numbers]{natbib}
\usepackage{amsmath}
\usepackage{bm}
\usepackage{pifont}
\newcommand{\bfvspace}[1]{{\vspace{0.5mm}\noindent{\textbf{#1}}}}

\def\tsc#1{\csdef{#1}{\textsc{\lowercase{#1}}\xspace}}
\tsc{WGM}
\tsc{QE}
\tsc{EP}
\tsc{PMS}
\tsc{BEC}
\tsc{DE}
\newcommand{\name}[0]{{ReL-NWM}}

\begin{document}
\let\WriteBookmarks\relax
\def\floatpagepagefraction{1}
\def\textpagefraction{.001}
\shorttitle{ReL-NWM}
\shortauthors{Zhiwei Zhang et~al.}

\title [mode = title]{Efficient Image-Goal Navigation with Representative Latent World Model}                      


\author[1]{Zhiwei Zhang}[style=chinese,orcid=0009-0008-8960-6223]
\ead{zhangzhiweizzw@nudt.edu.cn}
\credit{Writing - original draft, Validation,  Methodology , Formal analysis, Conceptualization}

\author[1]{Hui Zhang}[style=chinese,orcid=0009-0009-5435-9602]
\ead{huizhang_nudt@nudt.edu.cn}
\credit{Supervision, Funding acquisition}

\author[1]{Kaihong Huang}[style=chinese,orcid=0000-0002-2424-8524]
\credit{Writing - review and editing,Formal analysis}
\ead{kaihong.huang@nudt.edu.cn}

\author[1]{ Huimin Lu}[style=chinese,orcid=0000-0002-6375-581X]
\ead{lhmnew@nudt.edu.cn}
\credit{Supervision, Funding acquisition}

\author[1]{Chenghao Shi}[style=chinese,orcid=0000-0002-1462-2367]
\cormark[1]
\ead{shichenghao17@nudt.edu.cn}
\credit{Writing - review and editing,Funding acquisition,Validation,  Methodology , Formal analysis,Supervision}

\affiliation[1]{organization={ College of Intelligence Science and Technology, the National Key Laboratory of Equipment State Sensing and Smart Support, National University of Defense Technology},
                city={Changsha},
                postcode={410000}, 
                state={Hunan},
                country={China}}

\cortext[cor1]{Corresponding author}

\begin{abstract}
World models enable robots to conduct counterfactual reasoning in physical environments by predicting future world states. While conventional approaches often prioritize pixel-level reconstruction of future scenes, such detailed rendering is computationally intensive and unnecessary for planning tasks like navigation. We therefore propose that prediction and planning can be efficiently performed directly within a latent space of high-level semantic representations. To realize this, we introduce the Representative Latent space Navigation World Model~(\name{}). Rather than relying on reconstruction-oriented latent embeddings, our method leverages a pre-trained representation encoder, DINOv3, and incorporates specialized mechanisms to effectively integrate action signals and historical context within this representation space.
By operating entirely in the latent domain, our model bypasses expensive explicit reconstruction and achieves highly efficient navigation planning. Experiments show state-of-the-art trajectory prediction and image-goal navigation performance on multiple benchmarks. Additionally, we demonstrate real-world applicability by deploying the system on a Unitree G1 humanoid robot, confirming its efficiency and robustness in practical navigation scenarios.
\end{abstract}




\maketitle

\section{Introduction}

In autonomous navigation, conventional approaches typically follow modular pipelines that separate localization, mapping, and planning~\cite{10015689,yang2022far,cao2021exploring}. However, the inherent rigidity and susceptibility to error accumulation in such decoupled systems have motivated a shift toward end-to-end learning frameworks~\cite{fgprompt2023,zeng2024poliformer}. Among these, world model–based approaches~\cite{bar2025navigation,wang2023dreamwalker,koh2021pathdreamer,liu2024x} have gained prominence for enabling agents to perform counterfactual reasoning about future states through internal world simulation, mirroring the capacity for mental imagery in humans.


Despite these advances, constructing a navigation world model for real-world deployment remains challenging. Existing methods~\cite{bar2025navigation} typically adopt a standard strategy that utilizes a VAE encoder~\cite{blattmann2023stable} for pixel encoding and reconstruction, combined with a Diffusion Transformer (DiT)~\cite{bar2025navigation} for future state prediction. However, conventional VAE encoders compress visual inputs into low-dimensional latent spaces that limit information capacity, yielding weak representations due to their purely reconstruction-based training objective. Furthermore, while DiT enable fine-grained pixel-wise reconstruction, they require substantial computational resources for iterative denoising, which is prohibitively expensive and unnecessary for navigation tasks. Recent approaches~\cite{zhang2025latent,zhou2024dino} have attempted to address DiT's computational cost by replacing it with transformer-based state transition models for prediction. Nevertheless, these methods remain constrained by their reliance on VAE-encoded latent spaces, which are still limited by their reconstruction-focused nature.

To overcome these limitations, we propose the use of DINOv3~\cite{simeoni2025dinov3} as our representation encoder. DINOv3 provides high-level, semantically structured representations that enable reasoning over meaningful navigation cues, such as scene geometry and obstacle layout, rather than low-level pixel statistics. This approach offers a more efficient and task-relevant representation for navigation planning while maintaining strong predictive performance.


Building on this representation embedding, we incorporate several specific mechanisms to enable more effective integration of action and historical information. We utilize a Feature-wise Linear Modulation (FiLM) mechanism~\cite{perez2018film} for action conditioning. We posit that an action functions not merely as an attribute but as a dynamic modulator that transforms the environment's state. FiLM operationalizes this by allowing control signals to multiplicatively scale and additively shift feature channels, ensuring that the predicted future is both physically consistent and tightly governed by the agent's intent.
Furthermore, to capture long-term motion trends, we employ a Spatiotemporal Cross-Attention mechanism~\cite{shen2025effonav} that actively queries relevant historical patterns from the entire context window during inference. 

We designed a series of experiments demonstrating that our method achieves state-of-the-art trajectory prediction and image-goal navigation performance while exhibiting significant speed advantages. Given its lightweight architecture, we deployed our approach on a Unitree G1 humanoid robot, confirming that \name{} runs efficiently online on onboard hardware (Jetson Orin) and maintains strong robustness against real-world challenges. In summary, the main contributions of this paper are as follows:

\begin{itemize}
\item We present \name{}, a lightweight navigation world model that leverages representation encoder DINOv3 to achieve robust, reconstruction-free end-to-end image-goal navigation.
\item We introduce a dynamics model combining FiLM for active action conditioning and Cross-Attention for historical motion retrieval, jointly enabling state-of-the-art prediction accuracy.
\item We demonstrate state-of-the-art image-goal navigation performance in both simulated and real-world environment, while maintaining high computational efficiency.
\end{itemize}


\section{Related Work}
We review the latest advances in the two research areas most closely related to our work: end-to-end visual navigation and world model-based navigation.

\bfvspace{End-to-end visual navigation.} 
Recent paradigms have shifted towards learning-based approaches that map visual observations directly to actions, leveraging advancements in imitation learning (IL)~\cite{sridhar2024nomad,shah2023vint} and reinforcement learning (RL)~\cite{mezghan2022memory,zeng2024poliformer,fgprompt2023}.  PoliFormer~\cite{zeng2024poliformer}, which utilizes large-scale RL to enhance policy adaptability and practicality. More recently, generative diffusion models have been adopted to capture multi-modal distributions in navigation tasks. For instance, NoMaD~\cite{sridhar2024nomad} employs a unified diffusion strategy for both goal-directed navigation and exploration, while NavDP~\cite{cai2025navdp} integrates diffusion-based trajectory generation with a value function for improved real-world performance. GNM~\cite{shah2022gnm} further demonstrates that training on large-scale navigation datasets from diverse robotic platforms enhances generalization to unseen environments. Despite their success, these end-to-end methods face inherent limitations. Diffusion-based policies often incur high computational latency, creating bottlenecks for high-frequency real-time control. Furthermore, as purely reactive systems, they typically lack an explicit understanding of environmental dynamics, which limits their generalization capability in complex, dynamic, or unstructured scenarios.

\bfvspace{World models for navigation.} The capability to simulate future scenarios, termed "World Modeling", has become a cornerstone of modern embodied intelligence~\cite{hu2025simulating,hafner2025mastering}. Generally, these models serve two pivotal roles~\cite{ding2024understanding}: learning internal representations of the environment and predicting future dynamics to guide planning. Effective navigation necessitates a structured understanding of the scene. Early attempts, such as Pathdreamer~\cite{koh2021pathdreamer} and DreamWalker~\cite{wang2023dreamwalker} generate high-fidelity 2D images or structured abstractions to aid spatial understanding. More recently, WMNav~\cite{nie2025wmnav} integrated Vision-Language Models (VLMs) to estimate target probabilities, enhancing semantic understanding yet treating dynamics only implicitly. Our work aligns with the model-based paradigm but seeks to bridge the gap between high-level semantic reasoning and low-level control dynamics.

While recent diffusion-based models, such as NWM~\cite{bar2025navigation}, can generate high-fidelity videos of future states, they incur prohibitive computational costs, rendering them unsuitable for real-time control loops. Furthermore, these methods typically rely on reconstruction-based encoders or generative decoding, which can be sensitive to visual noise. In contrast, our approach leverages the pre-trained robustness of Foundation Models (specifically DINOv3)~\cite{simeoni2025dinov3} to construct a latent representation that provides spatially consistent and semantically rich features. Moreover, prior latent dynamics models often utilize simple concatenation to fuse actions with states, failing to capture the complex causal influence of control signals,such as DINO-WM~\cite{zhou2024dino} and LS-NWM~\cite{zhang2025latent}. This naive fusion often leads to predictions where the robot's intent is ignored. To address this, we introduce a Feature-wise Linear Modulation (FiLM)~\cite{perez2018film} mechanism to model transitions. Unlike passive concatenation, FiLM allows the action signal to multiplicatively modulate the feature channels of the current state. This enforces a strong structural conditioning, ensuring that the predicted future is strictly and physically governed by the agent's actions.

Our approach diverges from these methods by introducing a lightweight world model that operates entirely in a latent space. Instead of performing computationally expensive video generation, our method focuses on predicting transitions within implicit feature representations. A key feature of our architecture is the use of a DINOv3-based representation encoder for latent embedding, along with the integration of two complementary mechanisms that enable efficient fusion of action and historical information in this latent representation space.


\begin{figure*}  
  \centering
  \includegraphics[width=\linewidth]{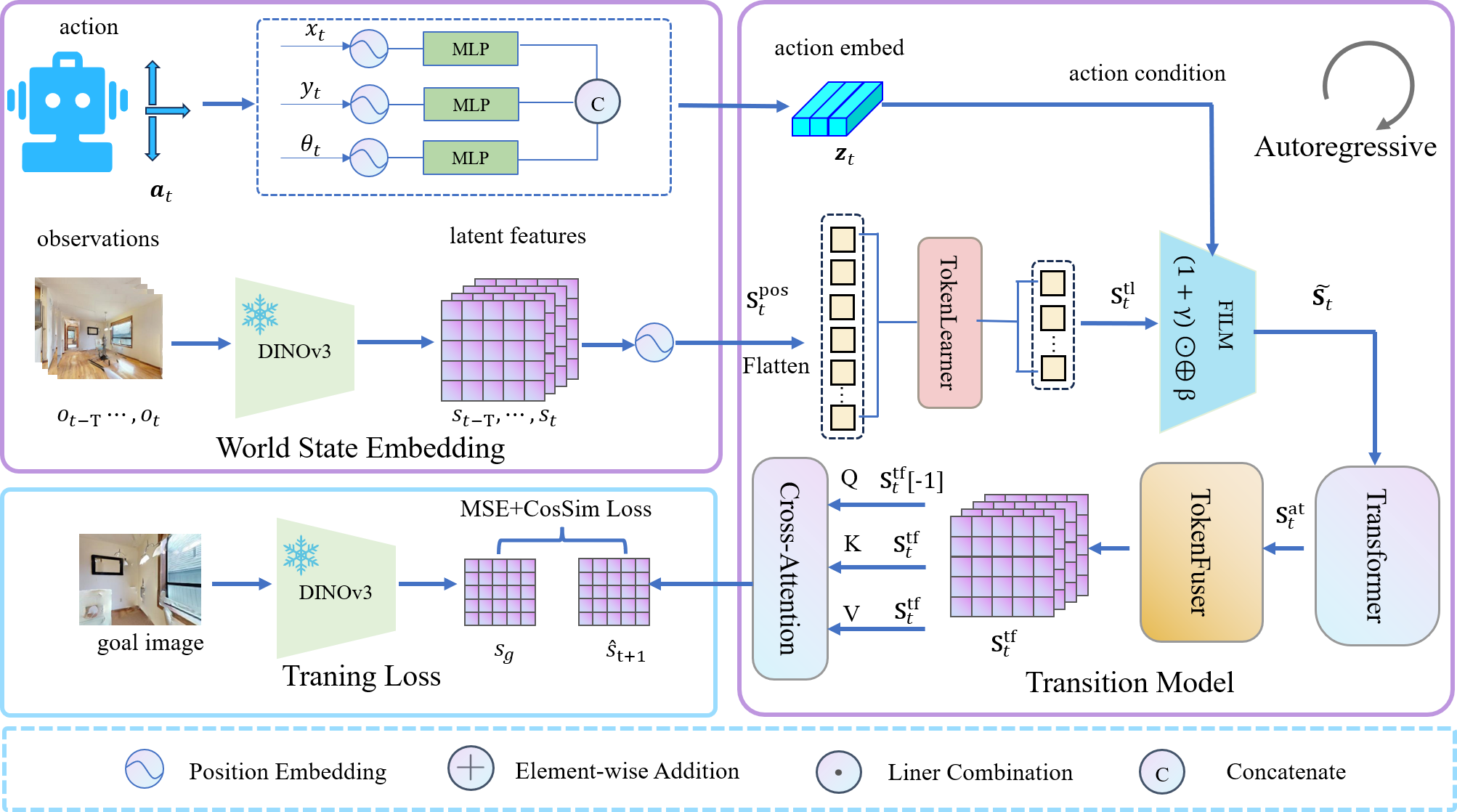} 
  \caption{Overview of the Representative Latent Navigation World Model (\name{}). Our world model comprises two core modules: a World State Embedding that encodes images and robot actions into latent state representations, and a Transition Model that predicts future latent states based on the embedded world state. The model is trained autoregressively using a combination of MSE and Cosine Similarity losses.}
  \label{fig:framework}
\end{figure*}

\section{Our Approach}

We introduce Representative Latent Navigation World Model~(\name{})  for end-to-end image goal navigation.  Given a sequence of visual observations $\mathcal{O}=\{\mathbf{o}_1,\mathbf{o}_2,\cdots,\mathbf{o}_n\}$ where $\mathbf{o}_t \in \mathbb{R}^{3\times H \times W}$,  and corresponding navigation actions $\mathcal{A}=\{\bm{a}_1,\bm{a}_2,\cdots,\bm{a}_n\}$ with $\bm{a}_t=(x_t,y_t,\theta_t)$ comprises forward translation $x_t$, lateral translation $y_t$, and rotation angle $\theta_t$, our model learns the representative latent dynamics of the environment directly from such video-like datasets.
This process is formalized as a state-transition model defined by:
\begin{equation}
    \mathbf{s}_{t+1} = p(\mathbf{s}_{\le t}, \mathbf{z}_{\le t}),
\end{equation}
where $\mathbf{s}_{t}=\text{enc}_\text{o}(\mathbf{o}_t)$ is the latent state derived from the observation $\mathbf{o}_t$ via the observation encoder $\text{enc}_\text{o}$ and  $\mathbf{z}_{t}=\text{enc}_\text{a}(\bm{a}_t)$ is the latent state derived from the action $\mathbf{a}_t$ via the action encoder $\text{enc}_\text{a}$, as detailed in Sec \ref{sec:stateembeeding}. $\mathbf{s}_{\le t}=(\mathbf{s}_1, \mathbf{s}_2, \dots, \mathbf{s}_t)$ and $\mathbf{z}_{\le t}=(\mathbf{z}_1, \mathbf{z}_2, \dots, \mathbf{z}_t)$ denote the sequences of encoded observations and actions, respectively. Further details regarding the transition model architecture and training methodology are provided in Sec \ref{sec:trasition} and  Sec \ref{sec:training}. 
Therefore, we forgo the computationally expensive decoder and perform planning directly in the latent space, enabling end-to-end image-goal navigation, as detailed in Sec \ref{sec:plan}.

\subsection{Representative Latent State Embedding}
\label{sec:stateembeeding}
The choice of latent representation space is critical for future state prediction and generation. In generative world models, VAE encoders~\cite{blattmann2023stable}, which prioritize pixel-level reconstruction, are generally considered more suitable for video generation than encoders like DINO~\cite{simeoni2025dinov3} or SigLIP~\cite{tschannen2025siglip} that focus on high-level semantics.
However, we hold that for navigation tasks, high-level semantic features are more relevant than fine-grained reconstruction. Additionally, recent advance, RAE~\cite{zheng2025diffusion}, challenges the conventional preference for VAEs by demonstrating that frozen representation encoders can serve as strong encoders for video generation. Motivated by these findings, we employ a pretrained DINOv3 ViT-S/16 encoder~\cite{simeoni2025dinov3} as our observation encoder to project inputs into a representative latent space. The encoded features are further projected into patch-wise representations  $\mathbf{s}_t \in \mathbb{R}^{M \times D}$, given as:
\begin{align}
  \mathbf{s}_t = \text{enc}_o(\mathbf{o}_t), 
\end{align}
where $M$ refers to the token number, and $D$ refers to the token dimension. A historical window of length $T$ is represented as $\mathbf{S}_t = (\mathbf{s}_{t-T+1}, \dots, \mathbf{s}_t) \in \mathbb{R}^{T \times M \times D}$. We flatten the temporal and spatial dimensions to form a unified spatiotemporal sequence $\mathbf{S}_t \in \mathbb{R}^{L \times D}$, where $L = T \times M$. To preserve the structural coherence of this flattened sequence, we inject learnable spatiotemporal positional information. Let $\mathbf{P}_\text{time} $ and $\mathbf{P}_\text{space}$ be the learnable temporal and spatial embeddings, respectively, the position-aware sequence is computed as:
\begin{equation}
    \mathbf{S}_t^\text{pos} = \text{LayerNorm}(\mathbf{S}_t + \mathbf{P}_\text{time} \oplus \mathbf{P}_\text{space}),
\end{equation}
where  the operator $\oplus$ represents element-wise addition with broadcasting. 

We encode the action vector $ \bm{a}_t = (x_t, y_t, \theta_t)$ using sinusoidal positional embeddings:
\begin{equation}
    \mathbf{z}_t = \text{Concat}\Big(\text{M}_x(\varepsilon(x_t)), \; \text{M}_y(\varepsilon(y_t)), \; \text{M}_{\theta}(\varepsilon(\theta_t))\Big),
\end{equation}
where $\varepsilon(\cdot)$ represents a standard sinusoidal frequency encoder that projects control signals into high-dimensional features, and $\text{M}_x$, $\text{M}_y$, and $\text{M}_{\theta}$ correspond to MLPs.

\subsection{Transition Model}
\label{sec:trasition}
Based on the encoded representative latent states, we further integrate actions with observations to facilitate action-conditioned prediction in the transition model.

To reduce computational complexity and support long-horizon planning, we employ a TokenLearner module~\cite{ryoo2021tokenlearner}. The TokenLearner takes $\mathbf{S}_t^\text{pos}$  as input and compresses the original $L$ tokens down to $K$ salient tokens $\mathbf{S}_t^\text{tl} \in \mathbb{R}^{K \times D}$  (where $K \ll L$), given as:
\begin{align}
  \mathbf{S}_t^\text{tl} = \mathrm{TokenLearner}( \mathbf{S}_t^\text{pos}).
\end{align}
These $K$ tokens form a compact summary of the spatiotemporal history, creating an information bottleneck that forces the model to distill only the most navigation-critical information in the latent space.



\bfvspace{FiLM-based State–Action Fusion.} A key goal of navigation world modeling is to make the latent representation responsive to the agent’s action. Instead of simple concatenation, we use Feature-wise Linear Modulation (FiLM)~\cite{perez2018film} to inject action information directly into the latent tokens. This provides a strong inductive bias that the future state is a transformation of the current state conditioned on the action. 
The formulation follows:
\begin{align}
  (\boldsymbol{\gamma}_t, \boldsymbol{\beta}_t)&=\mathrm{Linear}(\mathbf{z}_t),\\
  \tilde{\mathbf{S}}_t &= (1 + \boldsymbol{\gamma}_t) \odot \mathbf{S}_t^\text{tl} + \boldsymbol{\beta}_t
\end{align}
where $\boldsymbol{\gamma}_t, \boldsymbol{\beta}_t\in \mathbb{R}^{D}$ are scale and shift features projected from $\mathbf{z}_t$, respectively, and $\odot$ denotes element-wise multiplication.
FiLM provides continuous, dimension-wise modulation, enabling strong action–perception coupling while preserving lightweight computational cost. This also eliminates the bottleneck of scalar-channel gating and enables richer control over spatial latent tokens.
The features are then enhanced by Transformer, given as:
\begin{equation}
    \mathbf{S}_t^\text{at} = \text{Transformer}(\tilde{\mathbf{S}_t}).
\end{equation}

\bfvspace{Next-State Prediction.}
Before  generate the prediction for the next time step, we propagate the action-conditioned state from the $K$ tokens back to the original spatiotemporal layout $L$  using {TokenFuser},  we utilize the original position-aware sequence $\mathbf{S}_t^\text{pos}$ as a query to dynamically compute mixing weights, which are then used to re-distribute the compact token features back to their corresponding spatial coordinates, given as:
\begin{align}
  \mathbf{S}_t^\text{tf} = \mathrm{TokenFuser}( \mathbf{S}_t^\text{pos}, \mathbf{S}_t^\text{at}).
\end{align}
To generate the prediction for the next time step, we synthesize information from the full spatiotemporal context. Specifically, we extract the feature patches corresponding to the most recent observation and utilize them as queries. The entire fused history sequence serves as the keys and values in a SpatioTemporal Cross-Attention mechanism. The resulting context-aware features $\mathbf{u}_t$ are then passed through a residual MLP to predict the latent state of the next frame $\hat{\mathbf{s}}_{t+1}$, given as:
\begin{align}
  \hat{\mathbf{s}}_{t+1} = \mathrm{MLP}\big( \mathbf{u}_t \big) .
\end{align}

\subsection{Training Policy}
\label{sec:training}
Our world model is trained to predict future implicit states based on the current world state. In contrast to pixel-level supervision of future states, we directly supervise the predicted sparse implicit representations during training. Since navigation is inherently a long-horizon planning problem, accurate long-horizon predictions are crucial. To enhance long-term predictive consistency, we employ multi-step supervision, guiding the model to generate coherent future states over extended periods.

Specifically, we iteratively uses the predicted output along with ground-truth actions as input for subsequent predictions. This process continues for $N$ steps, with supervision applied at each step.
We optimize the model using a mean squared error (MSE) loss function to minimize the discrepancy between predicted and ground-truth latent features $\mathbf{s}_{t+k}^i$, formulated as:
    \begin{equation}
        \mathcal{L}_{mse}^{(k)} = \frac{1}{M} \sum_{i=1}^{M} \|\hat{\mathbf{s}}_{t+k}^i - \mathbf{s}_{t+k}^i\|_2^2,
    \end{equation}
    where
\begin{align}
\hat{\mathbf{s}}_{t+1} &= p_\theta(\mathbf{S}_t,\mathbf{z}_t), \qquad  \\
\hat{\mathbf{s}}_{t+k+1} &= p_\theta(\hat{\mathbf{S}}_{t+k}, \mathbf{z}_{t+k}), \;\; k \ge 1 ,
\end{align}
and { $\hat{\mathbf{S}}_{t+k}=(\mathbf{s}_{t+k-T+1},\cdots,\mathbf{s}_{t-1},\mathbf{s}_t,\hat{\mathbf{s}}_{t+1},\cdots,\hat{\mathbf{s}}_{t+k})$}, $p_\theta$ denotes our overall transition model, and $\theta$ represents the learnable model parameters. This autoregressive training mechanism enables the model to capture temporal dependencies across multiple steps, thereby improving the accuracy of long-horizon predictions.

Since DINOv3 features encode semantic similarity primarily through vector direction, relying solely on MSE can be insufficient. Therefore, we introduce a composite loss function combining Cosine Similarity Loss, given as:
    \begin{equation}
        \mathcal{L}_{cos}^{(k)} = 1 - \frac{1}{M} \sum_{i=1}^{M} \frac{\hat{\mathbf{s}}_{t+k}^i \cdot \mathbf{s}_{t+k}^i}{\|\hat{\mathbf{s}}_{t+k}^i\|_2 \|\mathbf{s}_{t+k}^i\|_2},
    \end{equation}
where $\|\cdot\|_2$ represents the $L_2$ norm.
The total loss for a trajectory of length $N$ is the weighted sum over all steps:
\begin{equation}
    \mathcal{L}_{total} = \frac{1}{N} \sum_{k=1}^{N} \left( \mathcal{L}_{cos}^{(k)} + \lambda \cdot \mathcal{L}_{mse}^{(k)} \right),
\end{equation}
where $\lambda$ is a balancing hyperparameter. 

\subsection{Image-goal navigation with \name{}}
\label{sec:plan}
In the image-goal navigation task, agents are required to navigate to a target location represented by a goal image. Following~\cite{bar2025navigation}, we formulate the navigation process as a Model Predictive Control (MPC) problem.

Leveraging our trained world model, we predict a sequence of future latent states from the initial state for each candidate action sequence. Instead of reconstructing future pixel observations, we optimize the action sequence $(\bm{a}_0, \dots, \bm{a}_{t-1})$ by minimizing the energy cost between the predicted final latent state $\hat{\mathbf{s}}_t$ and the goal latent state ${\mathbf{s}}_g$, where the energy cost is defined as:
\begin{equation}
    E(\hat{\mathbf{s}}_{t}, \mathbf{s}_g) = 1 - \text{CosSim}(\hat{\mathbf{s}}_t, \mathbf{s}_g).
\end{equation}
$\text{CosSim}(\cdot, \cdot)$ denotes the cosine similarity function, given as:
\begin{equation}
    \text{CosSim}(\hat{\mathbf{s}}_t^i, \mathbf{s}_g^i) = \frac{1}{M} \sum_{i=1}^{M}\frac{\hat{\mathbf{s}}_t^i \cdot {\mathbf{s}}_g^i}{\|\hat{\mathbf{s}}_t^i\|_2 \|\mathbf{s}_g^i\|_2},
\end{equation}
where $\mathbf{s}_t^i$ and $\mathbf{s}_g^i$ refers to $i$-th token of $\mathbf{s}_t$ and $\mathbf{s}_g$ respectively.

\begin{table*}[cols=8,pos=htbp] 
\caption{Trajectory Prediction Performance. ATE and RPE results across all in-domain datasets for trajectories predicted up to 2 seconds.}
\label{tab:result}
\begin{tabular*}{\tblwidth}{@{} LL *{6}{L} @{}}  
\toprule
\multirow{2}{*}{model} & \multirow{2}{*}{Encoder} & \multicolumn{2}{c}{RECON} & \multicolumn{2}{c}{HuRoN} & \multicolumn{2}{c}{Tartan} \\  
\cmidrule(lr){3-4} \cmidrule(lr){5-6} \cmidrule(lr){7-8}  
& & ATE & RPE & ATE & RPE & ATE & RPE \\  
\midrule
GNM       & CNN   & $1.87$       & $0.73$       & $3.71$ & $1.00$       & $6.65$       & $1.62$       \\  
NoMaD     & EfficientNet  & $1.95$       & $0.53$       & $3.73$       & $0.96$       & $6.32$       & $1.31$       \\  
NWM       & VAE   & $1.13$ & $0.35$ & $4.12$       & $0.96$       & $5.63$       & $1.18$       \\  %
LS-NWM    & VAE   & $1.51$       & $0.43$       & $4.54$       & $0.94$ & $\textbf{5.35}$ & $\textbf{1.13}$ \\  
\midrule
Ours      & DINOv3  & $\textbf{1.09}$       & $\textbf{0.30}$       & $\textbf{3.59}$       & $\textbf{0.93}$ & $5.57$ & $1.27$ \\ 
\bottomrule
\addlinespace[0.5em]
\multicolumn{8}{l}{\small The best results are highlighted in bold.}  
\end{tabular*}
\end{table*}

\section{Experiment Valuation}

We conducted experiments to evaluate the effectiveness of the proposed \name{}. First, we assessed the model’s trajectory prediction performance on standard benchmarks, as detailed in Sec \ref{sec:datasets}. To better understand the source of performance improvements, we performed ablation studies to analyze the contributions of key components, as presented in Sec \ref{sec:ablation}. Finally, we evaluated the image-goal navigation performance of our method in both simulated environments and a real-world Unitree G1 humanoid robot, as detailed in Sec \ref{sec:exp_n} and Sec \ref{sec:realrobot}, respectively.


\subsection{Experiments on Datasets}
\label{sec:datasets}
To quantitatively evaluate the predictive capability of our proposed method, we first conduct trajectory prediction experiments in the context of image-goal visual navigation. We compare our approach against several state-of-the-art baselines, including: end-to-end visual navigation models: GNM~\cite{shah2022gnm} and NoMaD~\cite{sridhar2024nomad}; and generative world model-based approaches: NWM~\cite{bar2025navigation} (using the default setting with 250 diffusion steps) and LS-NWM~\cite{zhang2025latent}.

\bfvspace{Datasets}. We utilize a diverse compilation of public datasets comprising first-person RGB images and odometry-derived poses:  RECON~\cite{shah2021rapid}: Off-road trajectories in unstructured outdoor terrains;
TartanDrive~\cite{triest2022tartandrive}: Aggressive driving data in challenging off-road environments; HuRoN~\cite{hirose2023sacson}: Crowded environments focusing on human-robot interaction.

\bfvspace{Evaluation Metrics}. We employ two standard metrics to assess prediction fidelity: Absolute Trajectory Error (ATE): Measures global consistency by computing the Euclidean distance between the predicted trajectory and the ground truth over the prediction horizon. Relative Pose Error (RPE): Evaluates local dynamics accuracy by measuring the relative pose discrepancy between consecutive time steps. 

\bfvspace{Results}. Table \ref{tab:result} presents the quantitative comparison of trajectory prediction performance over a 2-second horizon. Our method achieves state-of-the-art results on the challenging RECON and HuRoN datasets, significantly outperforming existing baselines.
While existing generative world models typically rely on a VAE encoder, they often capture high-frequency details that are irrelevant to control tasks. In contrast, our approach leverages the DINOv3 encoder, which prioritizes high-level semantic information over low-level pixel fidelity. This semantic-aware design enables our model to generalize more effectively in diverse and unstructured environments, where reconstruction-based methods tend to struggle.
The substantial performance gains also validate the effectiveness of the introduced FiLM mechanism and Spatiotemporal Cross-Attention module in capturing complex dynamics.

\subsection{Ablation Study on Datasets}
\label{sec:ablation}
To substantiate the design choices of our \name{}, we conduct ablation studies to isolate the individual contributions of its three core components: the Vision Encoder, the FiLM-based dynamics module, and the Spatiotemporal Cross-Attention mechanism. Specifically, we evaluate the impact of (i) replacing the DINOv3 encoder with a VAE, (ii) replacing the FiLM module with a passive concatenation of actions, and (iii) replacing the Spatiotemporal Cross-Attention with a CNN block. All variants were trained on the RECON dataset and evaluated using the Absolute Trajectory Error (ATE) and Relative Pose Error (RPE) metrics.

\bfvspace{Vision Encoder}. We first validate our choice of representation encoder by comparing our model against a VAE-based variant commonly used in reconstruction-based world models. As shown in Table~\ref{tab:ablation}, replacing DINOv3 with a VAE leads to a increase in prediction errors. This result aligns with our hypothesis: while VAEs excel at pixel-level reconstruction, they often encode high-frequency visual details that are irrelevant for control. In contrast, DINOv3 produces high-level semantic representations that capture structural and traversability information critical for navigation. 

\bfvspace{FiLM}. We posit that actions should serve as active operators that transform the state dynamics, rather than being treated as passive input features. To test this, we replace the FiLM module with a baseline that simply concatenates action embeddings to the latent state tokens. The results in Table~\ref{tab:ablation} show a significant performance drop when using concatenation, with ATE increasing by $63\%$ and RPE by $50\%$. This degradation confirms that FiLM results in sharper and more physically consistent dynamics.

\bfvspace{Spatiotemporal Cross-Attention}. The Spatiotemporal Cross-Attention module is designed to refine future predictions by selectively attending to relevant historical context. We ablate this component by replacing it with a standard CNN block. As reported in Table~\ref{tab:ablation}, this change causes the ATE increase by $50\%$. This finding underscores our design ensures better temporal consistency and trajectory accuracy.

\begin{table}[htbp]  
\centering
\caption{\textbf{Ablation Studies on RECON Dataset.} We evaluate the contribution of three key components: DINOv3, FiLM, and Cross-Attention. Best results are highlighted in \textbf{bold}.}
\label{tab:ablation}
\small  
\begin{tabular}{@{} c c c c c @{}}
\toprule
\textbf{FiLM} & \textbf{Cross-Attention} & \textbf{DINOv3} & \textbf{ATE $\downarrow$} & \textbf{RPE $\downarrow$} \\

\midrule
\ding{55} & \checkmark & \checkmark & 1.78 & 0.45 \\  
\checkmark & \ding{55} & \checkmark & 1.65 & 0.41 \\  
\checkmark & \checkmark & \ding{55} & 2.19 & 0.56 \\  
\textbf{\checkmark} & \textbf{\checkmark} & \textbf{\checkmark} & $\textbf{1.09}$ & $\textbf{0.30}$ \\  
\bottomrule
\addlinespace[0.3em]
\end{tabular}
\end{table}

\begin{table}[t]
\centering
\caption{\textbf{Navigation Performance \& Efficiency.} Comparison of Success Rate (SR), Path Efficiency (SPL), and average inference time per planning step.  Best results are highlighted in \textbf{bold}.}
\label{tab:navresult}
\begin{tabular}{@{} Lc c c @{}}
\toprule
\textbf{Method} & \textbf{SR} $\uparrow$ & \textbf{SPL} $\uparrow$ & \textbf{Time/Step (s)} $\downarrow$ \\
\midrule
NWM             & $20.0\%$ & $20.0\%$ & $55.31$ \\
LS-NWM          & $55.0\%$ & $30.9\%$ & $5.21$ \\
\textbf{\name{} (Ours)} & $\textbf{60\%}$ & $\textbf{48.5\%}$  & $\textbf{1.49}$ \\
\bottomrule
\end{tabular}
\end{table}



\begin{figure}  
  \centering
  \includegraphics[width=\linewidth]{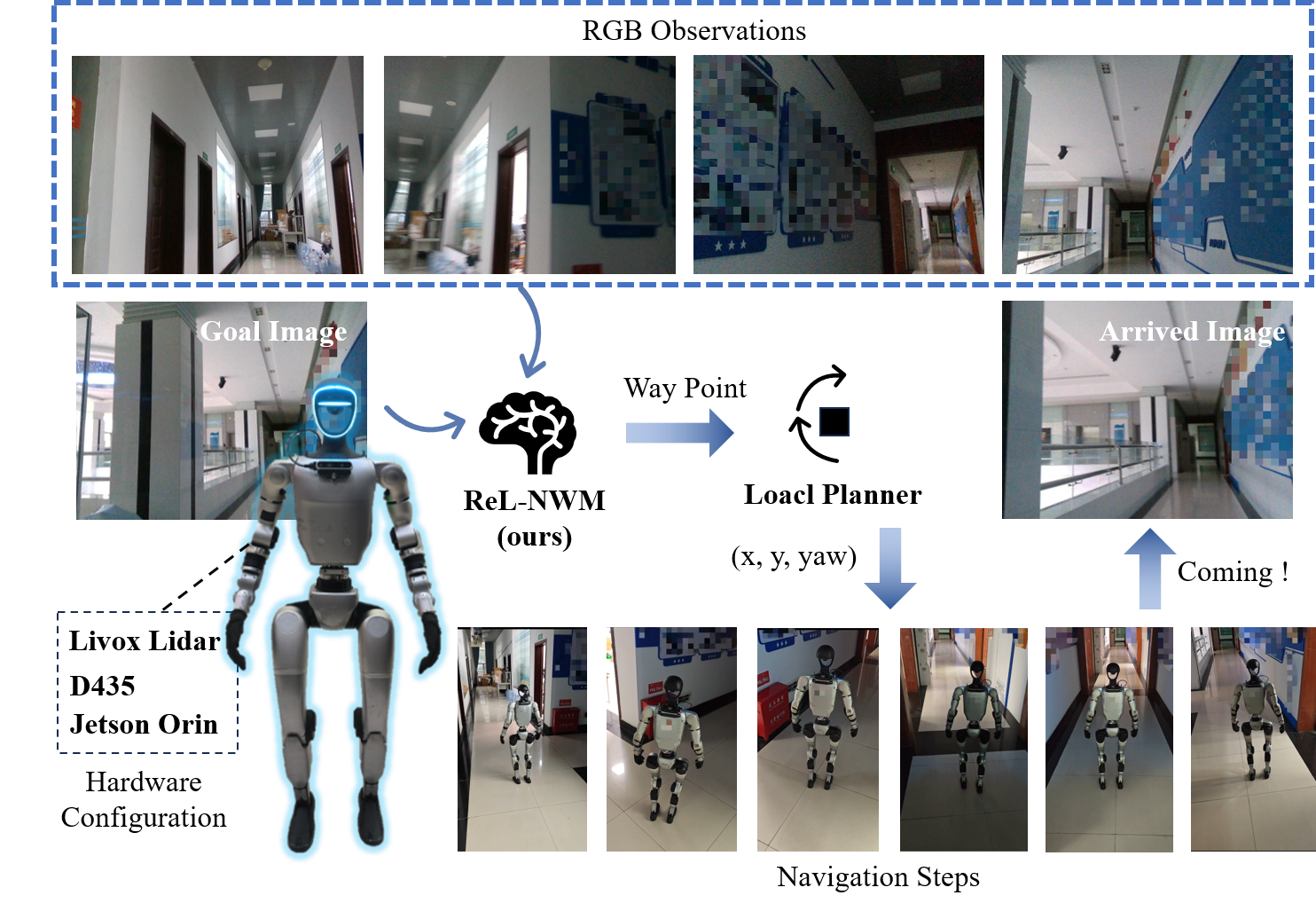} 
  \caption{Using the proposed \name{} on a Unitree G1, the robot navigates from a start position to a user-specified goal Image. The figure compares the robot's onboard view (top row) with its external motion (bottom row). The successful arrival (Arrived Image) highlights the model's robustness to real-world visual noise and its efficiency on onboard hardware.}
  \label{fig:robot}
\end{figure}

\subsection{Visual Navigation Performance}
\label{sec:exp_n}
Following~\cite{zhang2025latent}, we evaluate our approach on image-goal navigation tasks within the Habitat-Sim simulation environment, using the Gibson dataset, which offers diverse, high-fidelity photorealistic environments ranging from residences to offices.  comparing it against state-of-the-art baseline methods, including NWM and LS-NWM. A navigation episode is considered successful if the agent reaches within 2~m of the goal. The evaluation metrics include Success Rate (SR), Success weighted by Path Length (SPL), and the average time per planning step.

As shown in Table~\ref{tab:navresult}, our method achieves superior performance in both SR and SPL compared to the baseline methods. Notably, the most significant advantage of our approach lies in its computational efficiency. While NWM requires approximately 55~s to generate a single planning step, making it unsuitable for online control, the lighter LS-NWM still incurs a latency of 5.21~s due to the overhead of VAE decoding. In contrast, our method reduces the planning time to just 1.49~s per step. This represents an approximately 3× speedup over LS-NWM and a 36× speedup over NWM.
These results validate the effectiveness of our framework, which not only significantly improves computational efficiency but also enhances navigation performance

\subsection{Experiments on Real-World Humanoid Robot}
\label{sec:realrobot}

To further validate the generalization capability and real-world applicability of \name{} beyond simulation, we deployed our methods onto a physical robotic platform for indoor image-goal navigation. 

The experimental setup consists of a Unitree G1 humanoid robot equipped with a chest-mounted RGB camera as the only sensor. All computations were performed onboard using an NVIDIA Jetson Orin (16GB) module.
We implemented a hierarchical control framework to ensure safe and efficient operation. In this framework, \name{} serves as the high-level planner, generating waypoints through trajectory optimization using the Cross-Entropy Method (CEM). At each planning cycle, the system samples 120 candidate action sequences, performs 3 optimization iterations, and predicts future states over an 8-step horizon. Despite the computational complexity of model-based planning, our efficient latent-space architecture enables the system to publish optimized waypoints at approximately 1~Hz. These waypoints are subsequently processed by a low-level autonomy stack~\cite{zhang2020falco}, which generates smooth velocity commands and performs real-time obstacle avoidance. The navigation task requires the robot to traverse a long corridor and execute a precise turn at an intersection to reach the goal location. 
The navigation performance is quantitatively evaluated using the Success Rate (SR) metric.

\bfvspace{Results}.
Figure~\ref{fig:robot} illustrates a representative successful trial conducted in an office corridor environment. The results demonstrate that our method can effectively perform online planning to guide the robot to the destination without relying on prior maps or external localization systems. The robot successfully anticipates the corner, plans a smooth curved trajectory to navigate the intersection, and accurately arrives at the target location. This is evidenced by the high visual similarity between the robot's final observation and the goal image. Across 9 distinct trial scenarios, the robot successfully reached the goal destination 6 times, achieving a success rate of $66.7\%$. These results confirm that \name{} serves as a practical and efficient high-level planner capable of guiding physical agents through hierarchical control in real-world environments.


\section{Conclusion}
In this paper, we present \name{}, a lightweight framework for end-to-end image-goal navigation. Our approach bypasses computationally expensive pixel-level reconstruction and instead focuses on the high-dimensional semantic information most critical for navigation. Consequently, we replace computationally intensive VAE encoders with a representation encoder DINOv3. By incorporating FiLM for action conditioning and Spatiotemporal Cross-Attention for historical context modeling, our model achieves state-of-the-art dynamics prediction accuracy while maintaining significant efficiency advantages over existing baselines. We validate the proposed approach on image-goal navigation tasks in both simulated and real-world environments using a Unitree G1 humanoid robot. The results demonstrate that \name{} enables robust, online autonomous navigation on edge computing hardware.





\bibliographystyle{elsarticle-num-names}

\bibliography{new}


\end{document}